\documentclass[conference]{IEEEtran}
\IEEEoverridecommandlockouts
\usepackage{cite}
\usepackage{amsmath,amssymb,amsfonts}
\usepackage{algorithmic}
\usepackage{graphicx}
\usepackage{hyperref}
\usepackage{textcomp}
\usepackage{xcolor}
\usepackage{multirow}
\usepackage{booktabs}
\usepackage{siunitx}
\usepackage{rotating}
\def\BibTeX{{\rm B\kern-.05em{\sc i\kern-.025em b}\kern-.08em
    T\kern-.1667em\lower.7ex\hbox{E}\kern-.125emX}}
\begin{document}

\thispagestyle{empty}
\onecolumn
\vspace*{\fill}
\begin{center}
DOI: \href{https://doi.org/10.1109/IMCOM60618.2024.10418445}{10.1109/IMCOM60618.2024.10418445}
\end{center}
\vspace{1cm}
\begin{center}
\copyright 2024 IEEE. Personal use of this material is permitted. Permission from IEEE must be obtained for all other uses, in any current or future media, including reprinting/republishing this material for advertising or promotional purposes, creating new collective works, for resale or redistribution to servers or lists, or reuse of any copyrighted component of this work in other works.
\end{center}
\vspace*{\fill}
\clearpage

\twocolumn
\title{Efficient Ensemble for Multimodal Punctuation Restoration using Time-Delay Neural Network}

\author{\IEEEauthorblockN{Xing Yi Liu}
\IEEEauthorblockA{\textit{Department of Computer Science} \\
\textit{Columbia University}\\
New York, USA \\
liu.peter@columbia.edu}
\and
\IEEEauthorblockN{Homayoon Beigi}
\IEEEauthorblockA{\textit{Department of Computer Science} \\
\textit{Columbia University; Recognition Technologies, Inc.}\\
New York, USA; South Salem, USA \\
beigi@recotechnologies.com}
}

\IEEEoverridecommandlockouts
\IEEEpubid{\makebox[\columnwidth]{979-8-3503-3101-1/24/\$31.00 \copyright2024 IEEE \hfill}
\hspace{\columnsep}\makebox[\columnwidth]{ }}

\maketitle

\IEEEpubidadjcol

\begin{abstract}
Punctuation restoration plays an essential role in the post-processing procedure of automatic speech recognition, but model efficiency is a key requirement for this task. To that end, we present EfficientPunct, an ensemble method with a multimodal time-delay neural network that outperforms the current best model by 1.0 F1 points, using less than a tenth of its inference network parameters. We streamline a speech recognizer to efficiently output hidden layer acoustic embeddings for punctuation restoration, as well as BERT to extract meaningful text embeddings. By using forced alignment and temporal convolutions, we eliminate the need for attention-based fusion, greatly increasing computational efficiency and raising performance. EfficientPunct sets a new state of the art with an ensemble that weights BERT's purely language-based predictions slightly more than the multimodal network's predictions. Our code is available at \href{https://github.com/lxy-peter/EfficientPunct}{\texttt{https://github.com/lxy-peter/EfficientPunct}}.
\end{abstract}

\begin{IEEEkeywords}
speech recognition, punctuation restoration, multimodal learning, time-delay neural network, ensemble
\end{IEEEkeywords}

\section{Introduction}
\label{sec:intro}

Automatic speech recognition (ASR) systems' transformation of audio into text opens up possibilities for a wide range of downstream tasks. With natural language text, applications like  machine translation and voice assistants are enabled. However, raw ASR outputs lack punctuation and hence the full meaning of texts, which must be restored for usage by the aforementioned tasks. To illustrate the importance of punctuation, consider how the meaning of the sentence, ``I have a favorite, family," differs drastically from the unpunctuated version, ``I have a favorite family." Punctuation restoration is therefore also important for readability of transcribed speech and accuracy of conveyed message \cite{tundik2018user}.

Following the standard of the punctuation restoration task, we focus on three key punctuation marks which most commonly occur and play critical roles in language: commas (,), full stops (.), and question marks (?). We also consider no punctuation (NP) as a fourth class in need of our model's consideration.

\subsection{Related Work}
\label{ssec:related}

Many works and proposed architectures have been devoted to restoring punctuation, and two main research categories have emerged: (1) considering only text output from ASR, and (2) considering both text output from ASR and the original audio.

Most consider text only, effectively forming a natural language processing task. They usually train and evaluate on the benchmark, textual datasets from IWSLT 2011 and 2012. Researchers have studied a wide variety of methods, including $n$-gram models \cite{gravano2009restoring}, recurrent neural networks \cite{tilk2016bidirectional, kim2019deep, salloum2017deep}, adversarial models \cite{wang2022making}, contrastive learning \cite{huang2021token}, reinforcement learning \cite{lai2023boosting}, and transformers \cite{courtland2020efficient, alam2020punctuation, bakare2023punctuation, wang2023zephyr}. Conditional random fields \cite{lu2010better, nguyen2020improving, uyen2022vietnamese, ueffing2013improved} had particularly notable success. Direct fine-tuning of BERT \cite{kenton2019bert} has also proven effective, which we demonstrate in Section \ref{ssec:efficientpunct}.

In the other category, both audio and text modalities are considered. Earlier techniques involved statistical models like finite state machines \cite{christensen2001punctuation}, but unsurprisingly, more recently we see the exploration of neural networks \cite{tilk2015lstm, klejch2017sequence} and re-purposing existing models to take audio-based input and predict punctuation \cite{yi2019self, klejch2016punctuated}. Current state of the art models MuSe and UniPunc begin in separate branches: one to tokenize and process text and the other to process raw audio waveforms. They then use the attention mechanism \cite{bahdanau2015neural} to fuse text and acoustic embeddings \cite{sunkara2020multimodal, zhu2022unified}.

\subsection{Significance of Multimodal Approach}
\label{ssec:significance}

Despite research in multimodal punctuation restoration being far less numerous than the text-only category, \cite{yi2019self} explicitly demonstrated the value of added acoustic information. Intuitively, audio provides more diverse features from which models may learn \cite{beigi2011fundamentals}. As a simple example, long pauses in speech are definitive indicators of a full stop's (.) occurrence. Similarly, shorter pauses may indicate a comma (,), and rising pitch is often associated with question marks (?).

\begin{figure*}[t]
  \centering
  \includegraphics[width=17cm]{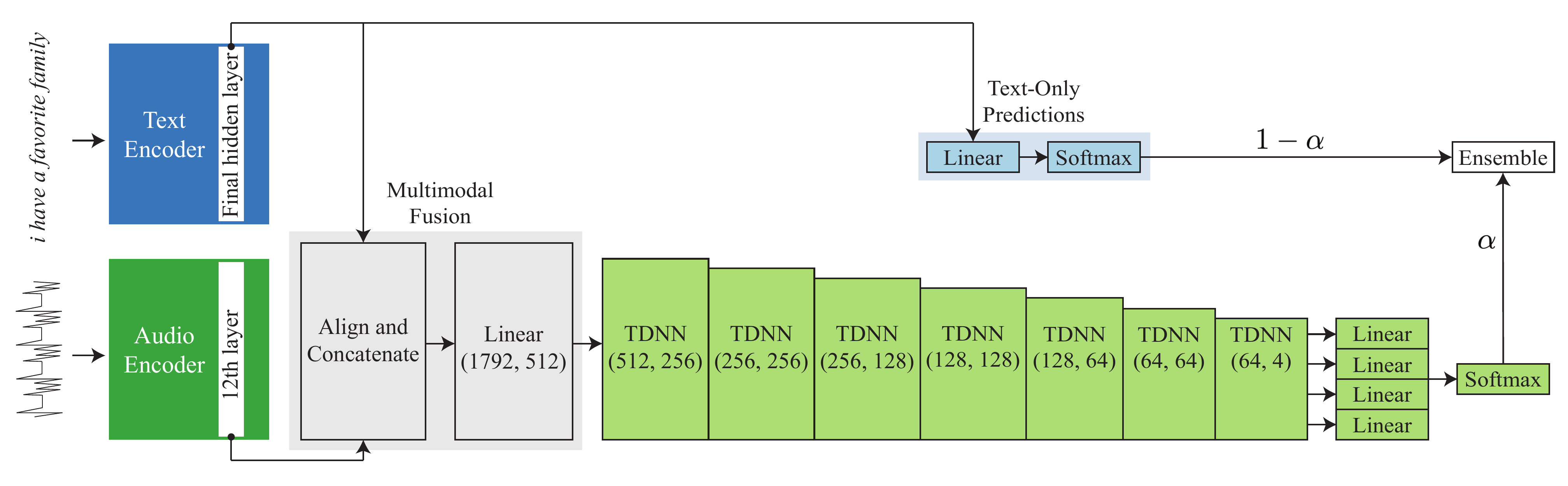}
  \caption{The EfficientPunct framework. The top branch predicts using text only, while the bottom branch predicts using text and audio.}
  \label{fig:overall}
\end{figure*}

The substantial benefit of involving both the transcribed text and original speech audio is that, in practical applications, we can design a highly streamlined system for restoring punctuation. Speech can first be transcribed into text by forward-passing audio signals through an ASR network, but one may preserve a hidden layer's latent representation for further usage as input (along with the transcribed text's embeddings) to a separate punctuation model. Then, the concatenated input would embed not only textual information, but also acoustics and prosody.

Our work is precisely motivated by this potential for high-speed punctuation labeling after receiving ASR output. We present EfficientPunct, a model that surpasses state of the art performance while requiring far fewer parameters, enabling practical usage.

\section{Method}
\label{sec:method}

We formulate the problem as follows. We are given spoken audio signal $\mathbf{a} = (a_1, a_2, \ldots, a_S)$ and transcription words $\mathbf{t} = (t_1, t_2, \ldots, t_W)$. Here, $S$ is the number of samples in the audio, and $W$ is the number of words. The goal is to predict punctuation labels $\mathbf{y} = (y_1, y_2, \ldots, y_W)$ that follow each word, where each $y_i \in \{\texttt{","}, \texttt{"."}, \texttt{"?"}, \texttt{NP}\}$. As illustrated in Fig. \ref{fig:overall}, EfficientPunct begins in two branches which separately process the audio signal $\mathbf{a}$ and transcription text $\mathbf{t}$. Their details are as follows.

\subsection{Text Encoder}
\label{ssec:text}

First, the text sequence $\mathbf{t}$ is passed through the default WordPiece tokenizer used by BERT. Then, using a pre-trained BERT model which we have fine-tuned for predicting the four aforementioned punctuation classes, we obtain final hidden layer text embeddings
\begin{equation}
H_t = \text{BERT}(\mathbf{t}).\label{eq:bert}
\end{equation}
$H_t$ is a matrix whose columns are $768$-dimensional vectors and represent embeddings of tokens. These text embeddings contain each token's context-aware information about grammar and linguistics.

\subsection{Audio Encoder}
\label{ssec:audio}

To process raw spoken audio waveforms and obtain meaningful acoustic embeddings, we use a pre-trained model built using the Kaldi speech recognition toolkit \cite{povey2011kaldi}. This is directly analogous to previous works' usage of wav2vec 2.0 \cite{baevski2020wav2vec} as their pre-trained audio encoder. Kaldi's TED-LIUM 3 \cite{hernandez2018tedlium} framework first extracts Mel frequency cepstral coefficients (MFCCs) \cite{beigi2011fundamentals} and i-vectors, which are then passed to a time-delay neural network for speech recognition. We extract the 12th layer's representation of the input audio for further usage in the punctuation model:
\begin{equation}
H_a = \text{KaldiTedlium12}(\mathbf{a}).\label{eq:tedlium}
\end{equation}
$H_a$ is a matrix whose columns are $1024$-dimensional embedding vectors. The number of columns is equal to the number of frames in the original audio.

\subsection{Alignment and Fusion}
\label{ssec:alignment}

The first step of fusing the $768$-dimensional embedding vectors from $H_t$ and the $1024$-dimensional embedding vectors from $H_a$ is to find correspondences between columns in each matrix. In other words, we must determine the text token being spoken during each frame of audio. This is performed through Kaldi's forced alignment algorithm. According to columns matched between the two modalities' embeddings, we concatenate them into columns of $1792$-dimensional embedding vectors. To fuse the two concatenated portions of each vector, we use a linear layer to learn affine transformations of embeddings which may be useful to punctuation restoration.

Many related works opt for attention-based fusion of the two modalities, but we found forced alignment and a simple linear layer to be the most parameter-efficient and competitive approach. Through experiments, we determined that more sophisticated fusion methods like cross-attention were counterproductive.

\subsection{Time-Delay Neural Network}
\label{ssec:time}

Next, the fused embeddings are passed through a time-delay neural network (TDNN) \cite{waibel1989phoneme}. It contains a series of 1D convolution layers to capture temporal properties of the features, with a gradually decreasing number of channels. At the last convolution layer, there are $4$ channels, with each one corresponding to a punctuation class. The channels are passed through two linear layers with weights and biases shared among the channels to output $4$ values for softmax activation.

\subsection{Ensemble Method}
\label{ssec:ensemble}

To complete EfficientPunct, we create an ensemble of the main TDNN and predictions using BERT's text embeddings only. We pre-trained BERT using the dark- and light-blue modules in Fig. \ref{fig:overall}, which can still be used at inference time to obtain a set of predictions that only consider text, grammar, and linguistics. The other set of predictions obtained from the TDNN consider both text and audio.

Let $\alpha \in [0,1]$ be the weight assigned to the TDNN's predictions and $1-\alpha$ be the weight assigned to  BERT's predictions. Our final predicted punctuation is
\begin{equation}
f(\mathbf{a}, \mathbf{t}, \alpha) = \arg\max \left[ \alpha y_a + (1-\alpha) y_t \right],\label{eq:ensemble}
\end{equation}
where $y_a$ is the TDNN's softmax values and $y_t$ is BERT's softmax values. If either the TDNN or BERT outputs a maximum class probability much lower than $1$, then the other model may help resolve the ambiguity in predicting a punctuation mark.

\section{Experiments}
\label{sec:experiments}

\subsection{Data}
\label{ssec:data}

Our primary dataset is the publicly available MuST-C version 1 \cite{cattoni2021mustc}, the same as that used by UniPunc \cite{zhu2022unified} for sake of fair comparison. This dataset was compiled using TED talks. We also use same training and test set splits as the original authors, whose information is available on GitHub. We further split the original training set into 90\% for training and 10\% for validation. Please see Table \ref{tab:split} for full information.

\begin{table}[t]
  \caption{Training, validation, and test set information}
  \label{tab:split}
  \centering
  \begin{tabular}{ c c c }
    \toprule
    \textbf{Set} & \textbf{Number of samples} & \textbf{Total duration (\SI{}{\hour})} \\
    \midrule
    Training & 92,723 & 392.0 \\
    Validation & 10,301 & 43.5 \\
    Test & 490 & 2.8 \\
    \bottomrule
  \end{tabular}
\end{table}

\begin{figure}[t]
  \centering
  \includegraphics[width=0.8\linewidth]{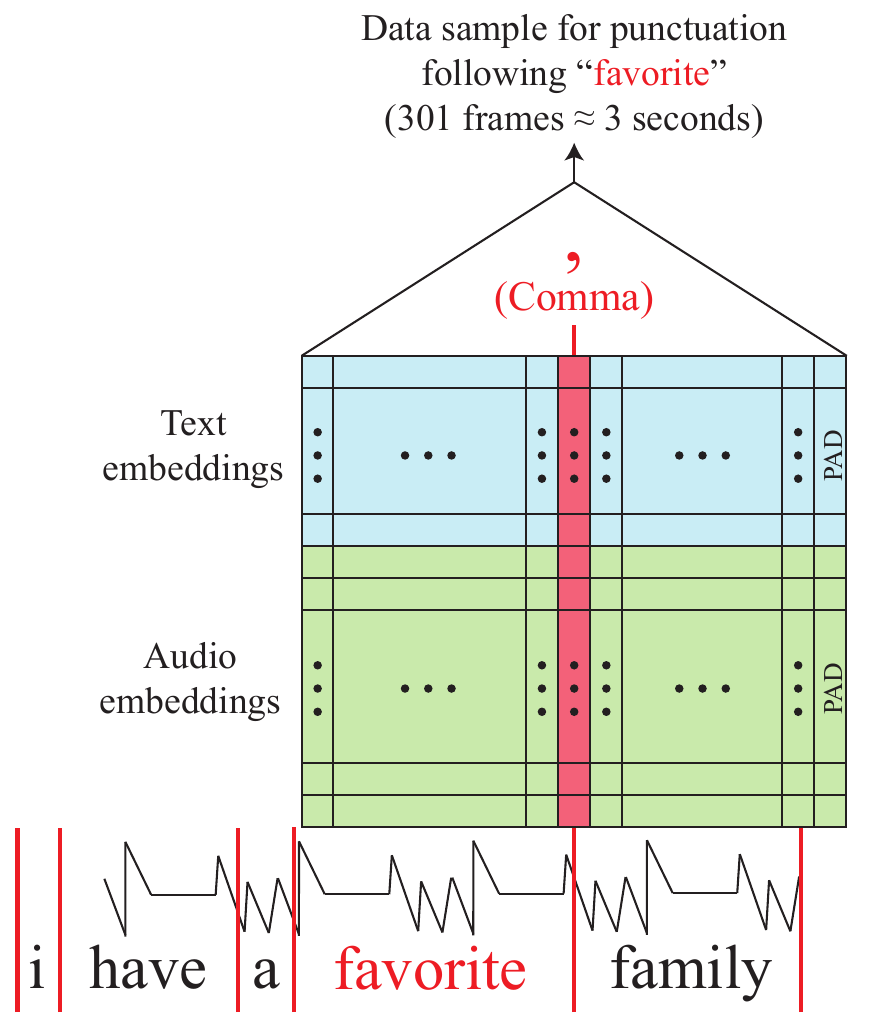}
  \caption{An example of preparing a data sample. We take 301 frames/columns, centered at the punctuation mark, from the matrix of concatenated text and audio embeddings.}
  \label{fig:data}
\end{figure}

Each sample is an English audio piece of approximately \SI{10}{\second} to \SI{30}{\second} with the corresponding transcription text. In Kaldi, we use a frame duration of \SI{10}{\milli\second} for MFCCs, i-vectors, and 12th layer acoustic embeddings. We follow the procedure described in Section \ref{ssec:alignment} to generate a matrix of aligned embeddings for each data sample. Then, to obtain data samples for training and inference, we consider segments of $301$ frames, or \SI{3}{\second}, wherein the exact middle frame is the point of transition from one text token to the next, as illustrated by the red column in Fig. \ref{fig:data}. The resulting data sample will thus be labeled with the punctuation following the prior token (``favorite" in Fig. \ref{fig:data}'s example) and occurring at the middle frame.

We use a context window of \SI{3}{\second}, because this duration should be sufficient to capture all acoustic and prosodic information relevant to a punctuation mark, such as pauses and pitch rises. At the same time, this duration is not so long as to include much unnecessary information, such as extensions into adjacent words.

For the entire dataset, punctuation label distributions were as described in Table \ref{tab:labels}. Due to the highly imbalanced nature of the dataset, we oversampled the three minority classes (comma, full stop, and question mark) for training. This ensures that the network avoids learning only the prior probability distribution. For each minority class, we sampled uniformly without replacement until the available pool of data samples emptied, and we repeated this process until the class count matched the majority class'.
\begin{table}[b]
  \caption{Punctuation label distributions}
  \label{tab:labels}
  \centering
  \begin{tabular}{ c c c }
    \toprule
    \textbf{Label} & \textbf{Number of samples} & \textbf{\% of total}\\
    \midrule
    No punctuation (NP) & 3,567,572 & 86.9\% \\
    Comma (,) & 280,446 & 6.8\% \\
    Full stop (.) & 238,213 & 5.8\% \\
    Question mark (?) & 20,897 & 0.5\% \\
    \bottomrule
  \end{tabular}
\end{table}

Moreover, since BERT was already pre-trained on massive corpora, we fine-tune it for punctuation prediction using the National Speech Corpus \cite{koh2019building} of Singaporean English, in addition to MuST-C.

\subsection{Training}
\label{ssec:training}

To fine-tune BERT and pre-train the text encoder, we place two linear layers on top of the base, uncased BERT's last hidden layer for four-way classification. For the pre-trained audio encoder, we use the TED-LIUM 3 \cite{hernandez2018tedlium} model in Kaldi.

\begin{table*}[t]
  \caption{F1 scores of EfficientPunct and its various submodules on each punctuation type, compared against existing state of the art (SOTA) models. EfficientPunct-BERT considers text only, EfficientPunct-TDNN considers text and audio, and EfficientPunct predicts using an ensemble of the prior two.}
  \label{tab:results}
  \centering
  \begin{tabular}{ c | c | c | c c c c | c }
    \toprule
    & \multirow{2}{*}{\textbf{Model}} & \textbf{Embedding} & \multirow{2}{*}{\textbf{Comma}} & \multirow{2}{*}{\textbf{Full Stop}} & \multirow{2}{*}{\textbf{Question}} & \multirow{2}{*}{\textbf{Overall}} & \textbf{Number of} \\
    & & \textbf{Type(s) Used} & & & & & \textbf{Parameters} \\
    \midrule
    \multirow{2}{*}{\begin{sideways}SOTA\end{sideways}} & MuSe$^\text{a,b}$ & BERT, wav2vec 2.0 & 73.2 & 83.6 & 79.4 & 77.9 & $1.7 \times 10^8$ \\
    & UniPunc$^\text{a}$ & BERT, wav2vec 2.0 & 74.2 & 83.7 & 80.8 & 78.5 & $2.5 \times 10^8$ \\
    \midrule
    \multirow{3}{*}{\begin{sideways}Ours\end{sideways}} & EfficientPunct-BERT & BERT & 73.4 & 83.9 & 84.7 & 78.4 & $1.1 \times 10^8$ \\
    & EfficientPunct-TDNN & BERT, TED-LIUM 3 & 74.3 & 83.6 & 85.8 & 78.5 & $1.2 \times 10^8$ \\
    & EfficientPunct (Ensemble) & BERT, TED-LIUM 3& \textbf{75.4} & \textbf{84.3} & \textbf{86.5} & \textbf{79.5} & $1.2 \times 10^8$ \\
    \bottomrule
    \multicolumn{8}{l}{$^\text{a}$Statistics taken directly from UniPunc paper due to public inaccessibility of certain models hindering our ability to run them.} \\
    \multicolumn{8}{l}{Fairness of comparison is ensured, since we use the exact same training and test sets as the UniPunc authors.} \\
    \multicolumn{8}{l}{$^\text{b}$Number of parameters in MuSe was conservatively estimated from information provided in the original paper.}
  \end{tabular}
\end{table*}

Our main TDNN module for punctuation restoration comprises seven 1-dimensional convolution layers, with said dimension spanning across time. Fig. \ref{fig:overall} shows the number of input and output channels of each layer. The kernel sizes used are, in order: $9$, $9$, $5$, $5$, $7$, $7$, $5$, alternating between no dilation and a dilation of $2$. The stride was kept at $1$ in all layers. Additionally, we apply ReLU activation and batch normalization to the output of each layer. We trained using stochastic gradient descent with learning rate $0.00001$ and momentum $0.9$, instead of the typically used Adam optimizer. This allowed for greater generalizability but still reasonable training speed \cite{gupta2021adam}.

To experiment with our ensemble, we explored the effect of varying $\alpha$, the weight assigned to the TDNN for final predictions. $1-\alpha$ is the weight assigned to BERT. In Section \ref{sec:results}, we report results for $\alpha=0.3$ to $\alpha=0.7$ in $0.1$ increments.

We used a standard Linux computing environment hosted on Google Cloud Platform with a single NVIDIA Tesla P100 GPU. Training took roughly 2 days, and inference can be performed on CPU-only machines 50 times faster than real time, or in about 0.02 seconds per second of audio.

\section{Results}
\label{sec:results}

Our results reported in Table \ref{tab:results} includes a comparison with current state of the art (SOTA) and best-performing models, MuSe \cite{sunkara2020multimodal} and UniPunc \cite{zhu2022unified}. We also divide the reporting of EfficientPunct's results into three categories:
\begin{enumerate}
\item \textit{EfficientPunct-BERT} considers text only, which is equivalent to the fine-tuned BERT model.
\item \textit{EfficientPunct-TDNN} considers text and audio via our TDNN.
\item \textit{EfficientPunct} is an ensemble of predictions from categories (1) and (2) with $\alpha=0.4$, the best performing weight as reported in Section \ref{ssec:weights}.
\end{enumerate}
Categories (1), (2), and (3) are reported in the third, fourth, and fifth rows of Table \ref{tab:results}, respectively.

As is standard in punctuation restoration research, we report the F1 scores of commas, full stops, and question marks. The ``overall" F1 score aggregates these while considering the imbalanced classes' varying numbers of data samples. We also state each model's number of parameters to provide an indication of computational efficiency.

\subsection{EfficientPunct and Submodules}
\label{ssec:efficientpunct}

Our main EfficientPunct model achieves an overall F1 score of 79.5, outperforming all current state of the art frameworks by 1.0 or more points. We also achieve highest F1 scores for each individual punctuation mark, with the most significant improvement occurring for question marks. These were accomplished with EfficientPunct using less than half of UniPunc's total number of parameters, which achieved the previous best results. The significant improvement in recognizing question marks may be attributed to our audio encoder, Kaldi's TED-LIUM 3 framework, aiming explicitly at phone recognition. In this process, the acoustics surrounding question marks may be more pronounced in the embedding representation than other acoustics models.

Even more lightweight models are EfficientPunct-BERT and EfficientPunct-TDNN. EfficientPunct-BERT is simply a concatenation of two linear layers and a softmax layer on top of BERT. With the incorporation of audio features, we observe that EfficientPunct-TDNN performs slightly better.

These results validate the strength of TDNNs in punctuation restoration, which are traditionally used in speech and speaker recognition. UniPunc and MuSe both used attention-based mechanisms for fusing text and acoustic embeddings, but alignments learned as such rely on trainable attention weights. Our forced alignment strategy likely generated more precise temporal matches between text and audio. Combined with a TDNN architecture, we achieved a significantly more efficient model.

\subsection{Ensemble Weights}
\label{ssec:weights}

In this section, we observe the effect of ensemble weights on EfficientPunct's performance. \eqref{eq:ensemble} details the role of $\alpha$ in weighting predictions made by the TDNN and BERT, with $\alpha=0$ meaning pure consideration of BERT, and $\alpha=1$ meaning pure consideration of the TDNN.

Table \ref{tab:alpha} reports the effect of $\alpha$ on model performance. When both BERT and the TDNN play an approximately equal role in the ensemble, a fair voting mechanism is enabled, and the highest F1 scores are achieved. However, notice that $\alpha=0.4$, a weight that considers BERT slightly more strongly than the TDNN, achieves the maximum overall F1. This gain comes mostly from sharper comma predictions, which present notorious difficulties due to varying grammatical and (transcription) writing styles. We reason that $\alpha=0.4$ excels, because a stronger reliance on BERT's language modeling perspective yields more linguistically correct punctuation, as agreed upon by countless writers' contributions to BERT's training corpora.

\begin{table}[t]
  \caption{F1 scores for different $\alpha$ weights}
  \label{tab:alpha}
  \centering
  \begin{tabular}{ c | c c c c }
    \toprule
    $\alpha$ & \textbf{Comma} & \textbf{Full Stop} & \textbf{Question} & \textbf{Overall} \\
    \midrule
    0.3 & 75.0 & 84.1 & 86.3 & 79.2 \\
    0.4 & 75.4 & 84.3 & 86.5 & \textbf{79.5} \\
    0.5 & 75.0 & 84.0 & 86.5 & 79.1 \\
    0.6 & 75.0 & 83.8 & 86.2 & 79.0 \\
    0.7 & 74.8 & 83.8 & 85.8 & 78.9 \\
    \bottomrule
  \end{tabular}
\end{table}

The strength of our ensemble method is that, in cases of uncertain predictions by either party, i.e. approximately equal softmax probabilities over all classes, the other can provide guidance to clarify the ambiguity. This process demands very little additional parameters through which the input must be passed, as shown by the last two rows of Table \ref{tab:results}, but greatly advances state of the art performance.

\subsection{Parameter Breakdown}
\label{ssec:parameter}

In order to show the specific modules in which we attain superior efficiency, we further break down the parameters count from the last column of Table \ref{tab:results}. In Table \ref{tab:parameters}, we detail the number of parameters devoted by each model to extracting embeddings and inferring them to make punctuation decisions.

\begin{table}[b]
  \caption{Number of parameters required in various stages of each model}
  \label{tab:parameters}
  \centering
  \begin{tabular}{ c | c c c }
    \toprule
    \multirow{2}{*}{\textbf{Model}} & \textbf{Embedding} & \textbf{Inference} & \multirow{2}{*}{\textbf{Total}} \\
    & \textbf{Network} & \textbf{Network} & \\
    \midrule
    MuSe & $1.6 \times 10^8$ & $4.3 \times 10^6$ & $1.7 \times 10^8$ \\
    UniPunc & $2.0 \times 10^8$ & $4.8 \times 10^7$ & $2.5 \times 10^8$ \\
    EfficientPunct & $\mathbf{1.1 \times 10^8}$ & $\mathbf{3.0 \times 10^6}$ & $\mathbf{1.2 \times 10^8}$ \\
    \bottomrule
  \end{tabular}
\end{table}

EfficientPunct requires much less computational cost in both the embedding extraction and inference stages. Our usage of Kaldi's TED-LIUM 3 model brought massive efficiency gains compared to MuSe and UniPunc's usage of wav2vec 2.0. Moreover, our inference module uses less than a tenth of UniPunc's parameters in the same stage, which achieved the previous best results.

\section{Conclusion}
\label{sec:conclusion}

In this paper, we explored the application of time-delay neural networks in punctuation restoration, which proved to be more computationally efficient than and as effective as previous approaches. Combined with BERT in an ensemble, EfficientPunct establishes a strong, new state of the art with a fraction of previous approaches' number of parameters. A key factor of our model's success is removing the need for attention-based fusion of text and audio features. In previous approaches, multiple attention heads added extraordinary overhead in the punctuation prediction stage. We demonstrated that forced alignment of text and acoustic embeddings, in conjunction with temporal convolutions, rendered attention unnecessary.

Additionally, we studied the effect of different weights assigned to members of the ensemble. We found that a slightly stronger weighting of BERT against the multimodal TDNN optimized performance by emphasizing language rules associated with punctuation.

In future works, the effectiveness of jointly training ensemble weights and the TDNN may be examined, as a current limitation is the rather simplistic, linear-weighted ensembling method. Jointly training with the text and audio encoders may also be considered, but this procedure should not inhibit the encoders' generalizability for purposes other than punctuation restoration. Finally, we would like to explore the applicability of EfficientPunct in more languages and a similar framework for other post-processing tasks of speech recognition.

\bibliographystyle{IEEEtran}

\end{document}